\documentclass{article}
\usepackage{graphicx} 
\usepackage{multirow}
\usepackage{float}
\usepackage{authblk}
\usepackage{amssymb,latexsym,amsfonts}

\begin{document}

\title{SAM for Poultry Science}

\renewcommand\Authsep{, }
\renewcommand\Authand{, }
\newcommand*\samethanks[1][\value{footnote}]
{\footnotemark[#1]}
\author[1]{Xiao Yang \thanks{Co-first authors}}
\author[2]{Haixing Dai \samethanks}
\author[2]{Zihao Wu}
\author[1]{Ramesh Bist}
\author[1]{Sachin Subedi}
\author[2]{Jin Sun}
\author[3]{Guoyu Lu}
\author[4]{Changying Li}
\author[2]{Tianming Liu \thanks{Co-corresponding authors. Email: tliu@uga.edu, lchai@uga.edu}}
\author[1]{Lilong Chai \samethanks}

\affil[1]{Department of Poultry Science, University of Georgia, Athens, 30602, USA}
\affil[2]{School of Computing, University of Georgia, Athens, 30602, USA}
\affil[3]{Department of Electrical \& Computer Engineering, University of Georgia, Athens, 30602, USA}
\affil[4]{Department of Agricultural and Biological Engineering, University of Florida, Gainesville, 32611, USA}

\maketitle
\begin{abstract}
\textbf{} \\
In recent years, the agricultural industry has witnessed significant advancements in artificial intelligence (AI), particularly with the development of large-scale foundational models. Among these foundation models, the Segment Anything Model (SAM), introduced by Meta AI Research, stands out as a groundbreaking solution for object segmentation tasks. While SAM has shown success in various agricultural applications, its potential in the poultry industry, specifically in the context of cage-free hens, remains relatively unexplored. This study aims to assess the zero-shot segmentation performance of SAM on representative chicken segmentation tasks, including part-based segmentation and the use of infrared thermal images, and to explore chicken-tracking tasks by using SAM as a segmentation tool. The results demonstrate SAM's superior performance compared to SegFormer and SETR in both whole and part-based chicken segmentation. SAM-based object tracking also provides valuable data on the behavior and movement patterns of broiler birds. The findings of this study contribute to a better understanding of SAM's potential in poultry science and lay the foundation for future advancements in chicken segmentation and tracking.
\end{abstract}

\vspace{12pt}


\section{Introduction}
In recent years, the field of artificial intelligence has made significant progress in the development of large-scale foundation models, which have had an impact on a variety of domains, including agriculture \cite{li2023learning, wang2021search, eli2019applications, lu2023agi}. These models, which incorporate natural language processing (NLP) and computer vision, have revolutionized the agricultural industry. Large Language Models (LLMs) have been instrumental in revolutionizing natural language processing in the agricultural domain. Notably, OpenAI models such as GPT4 have pushed the limits of language comprehension and generation in the context of agricultural data. These language models have demonstrated their zero-shot generalization capabilities and have applications in a variety of agricultural fields, including unmanned aerial vehicles (UAVs) management \cite{de2023semantic}, robotic gripper \cite{stella2023can} and weather forecasting \cite{vaghefi2023chatipcc}. The impressive achievements of large language models have inspired researchers to expand their investigation into the field of visual learning. Initially, the emphasis in visual foundation models was placed on pre-training methods that produce semantically rich numerical representations or linguistic descriptions of images. Recent emphasis has shifted to the creation of specialized foundation models for semantic segmentation tasks \cite{zhang2023self, luo2022multisource}. Such tasks require a thorough comprehension of the geometric structure of images, which necessitates accurate analysis and interpretation.

Recently, the Segment Anything Model (SAM) introduced by Meta AI Research is revolutionizing object segmentation tasks with its cutting-edge foundation model. Trained on an extensive dataset, SAM demonstrates remarkable zero-shot segmentation capabilities \cite{kirillov2023segment}. By supporting various visual prompts such as points, boxes, and masks, SAM offers versatility in image analysis, particularly in scenarios with limited labeled training data. SAM's unique prompt ability sets it apart from traditional segmentation models, making it an ideal solution for precise and efficient object segmentation. This breakthrough technology opens new possibilities for a range of applications, including agriculture. In the agricultural domain, where accurate object segmentation is crucial, SAM empowers researchers and practitioners to enhance pest and leaf disease detection and crop segmentation\cite{ji2023segment, tang2023can, chen2023sam}. However, the application of SAM in the poultry industry, particularly in the context of cage-free hens, remains limited. The growing trend in the egg industry towards cage-free housing systems, driven by the objective of enhancing bird welfare and complying with regulations such as the requirement for all eggs sold in California to come from cage-free houses, necessitates the development of automated methods for detecting laying hens on the litter floor \cite{subedi2023tracking, yang2022deep, yang2023deep, yang2023automatic,bist2023mislaying}. By leveraging the advanced capabilities of SAM, the poultry industry can significantly enhance precision and effectiveness in various analysis tasks.

This study focuses on evaluating the zero-shot segmentation performance of the SAM model on representative segmentation tasks and on exploring the utilization of SAM for object tracking. Specifically, we assess its ability to segment chickens under different scenarios, including part-based segmentation and the use of infrared thermal images. To evaluate SAM's performance, we compare the results with state-of-the-art (SOTA) domain-specific models. Additionally, we explore the potential of combining SAM with other multi-object tracking (MOT) methods. Through these investigations, we aim to enhance our understanding of SAM's capabilities and its effectiveness in addressing various challenges in chicken segmentation and tracking.

\section{Methods}
\textbf{Chicken segmentation.}  To evaluate the zero-shot segmentation performance of SAM on poultry imaging data, we utilized two distinct datasets of cage-free chickens, one consisting of source images and the other of thermal images. To assess the robustness of SAM, we conducted two chicken segmentation tasks: semantic segmentation and part-based segmentation (excluding the tails). Furthermore, we investigated two usage scenarios: (a) SAM with a single-point user prompt and (b) SAM with a total set of points provided for each object in a scene.

\noindent\textbf{Chicken tracking using SAM.} While SAM is not designed for the tracking task, we adapt it to a tracking workflow to perform this important task. A broiler chicken dataset was created to evaluate the performance of the tracking function. The dataset was designed specifically to test the ability of a tracking system to accurately follow the movements of broiler chickens. We propose a new combination of SAM and YOLOX\cite{ge2021yolox} + ByteTracker\cite{zhang2022bytetrack} as a powerful solution for tracking individual birds in images, which is a methodology contribution of this work. YOLOX is a real-time object detection and classification algorithm that uses a one-stage detector to predict both bounding boxes and class probabilities for objects in an image \cite{ge2021yolox, zhang2022bytetrack}. This information is used by ByteTracker to maintain a consistent track for each bird over time. SAM can extract the bounding box $(x, y, w, h)$ of each bird in an image, which provides corresponding locations for YOLOX + ByteTracker to accurately track the birds over time. 

\section{Results and discussion}
\subsection{Chicken segmentation}
Two state-of-the-art methods, SegFormer \cite{xie2021segformer} and SETR \cite{liu2022setr}, were used as a baseline for comparison with SAM. The results are presented in Table 1. SAM demonstrated superior performance compared to SegFormer and SETR in both the whole and part-based segmentation benchmarks. Especially when using the total points prompts, SAM had significant performance gains. We also found that source images were better suited for segmentation tasks compared to thermal images. This is because thermograms have various colors of pixels and the edge of the chicken does not have a clear boundary with the background, making it challenging for all models to accurately recognize chickens \cite{resendiz2017segmentation, zhang2023feather}. In terms of semantic segmentation and arbitrary parts, semantic segments performed better than arbitrary parts (excluding the tail) because the tail has a similar color to the chicken body under both natural light and thermal conditions, making it difficult to recognize the tail alone \cite{jing2023segment, chen2023learning, ahmadi2023application}. As a result, the recognition of the whole chicken body is an easier task.
\begin{figure}[H]
    \centering
    \includegraphics[width=\linewidth]{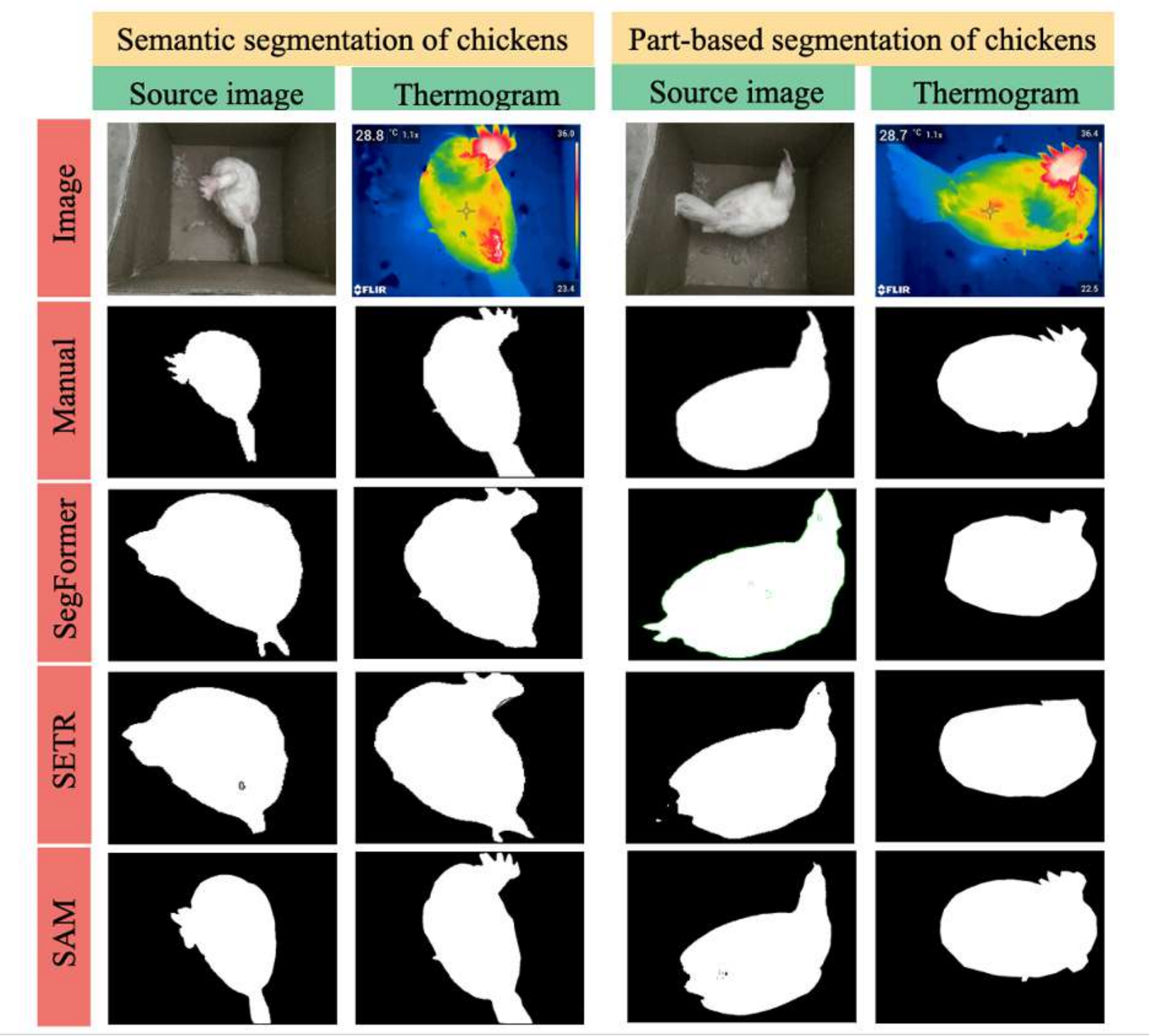}
    \caption{Visual comparison of the segmentation results. State-of-the-art methods (SegFormer and SETR) are compared with SAM applied to diverse chicken datasets.}
    \label{fig:seg}
\end{figure}
\begin{table}[htbp]
\centering
\caption{A comparison of SAM and state-of-the-art (SOTA) methods (SegFormer and SETR) in mean Intersection over Union (mIoU).}
\resizebox{\linewidth}{!}{
\begin{tabular}{lcccc}
\hline
\multicolumn{1}{c}{\multirow{2}{*}{Method}} & \multicolumn{2}{c}{Semantic segmentation of chickens} & \multicolumn{2}{c}{Part-based segmentation of chickens} \\ \cline{2-5} 
\multicolumn{1}{c}{}                        & Source image               & Thermogram               & Source image                & Thermogram                \\ \hline
SegFormer                                   & 43.22                      & 23.44                    & 35.34                       & 0.22                      \\
SETR                                        & 42.90                      & 35.34                    & 29.91                       & 0.28                      \\
SAM(one point)                              & 92.50                      & 88.95                    & \textbf{86.17}                       & 72.20                     \\
SAM(total points)                           & \textbf{94.80}                      & \textbf{91.74}                    & 85.64                       & \textbf{80.08}                     \\ \hline
\end{tabular}
}
\end{table}

\subsection{Track chickens using SAM}
The combination of SAM, YOLOX, and ByteTracker enables the real-time tracking of individual broiler bird movements in videos. As shown in Fig. 2a and 2b, the tracking system accurately tracks the movement of a chicken as it moves from the center of the image to the feeder. The initial bounding box information $(x, y, w, h)$ extracted by SAM is input into the YOLOX algorithm, which classifies the object within the bounding box as a broiler bird. The output of YOLOX, including the updated bounding box information, is then fed into the ByteTracker algorithm, which tracks the broiler bird as it moves through the scene (from Fig. 2a $(426.9, 368.0, 57.0, 95.0)$ to Figure 2b $(399.1, 472.0, 52,7, 96.6)$). The use of YOLOX and ByteTracker in this workflow enables the system to accurately track the movement of the broiler bird regardless of its various positions and orientations in the complex scene \cite{rong2023tomato, cao2023retinamot}. Our SAM-powered tracking result can provide valuable information regarding the behavior and movement patterns of broiler birds, which can be utilized to optimize broiler production operations.
\begin{figure}[htbp]
    \centering
    \includegraphics[width=\linewidth]{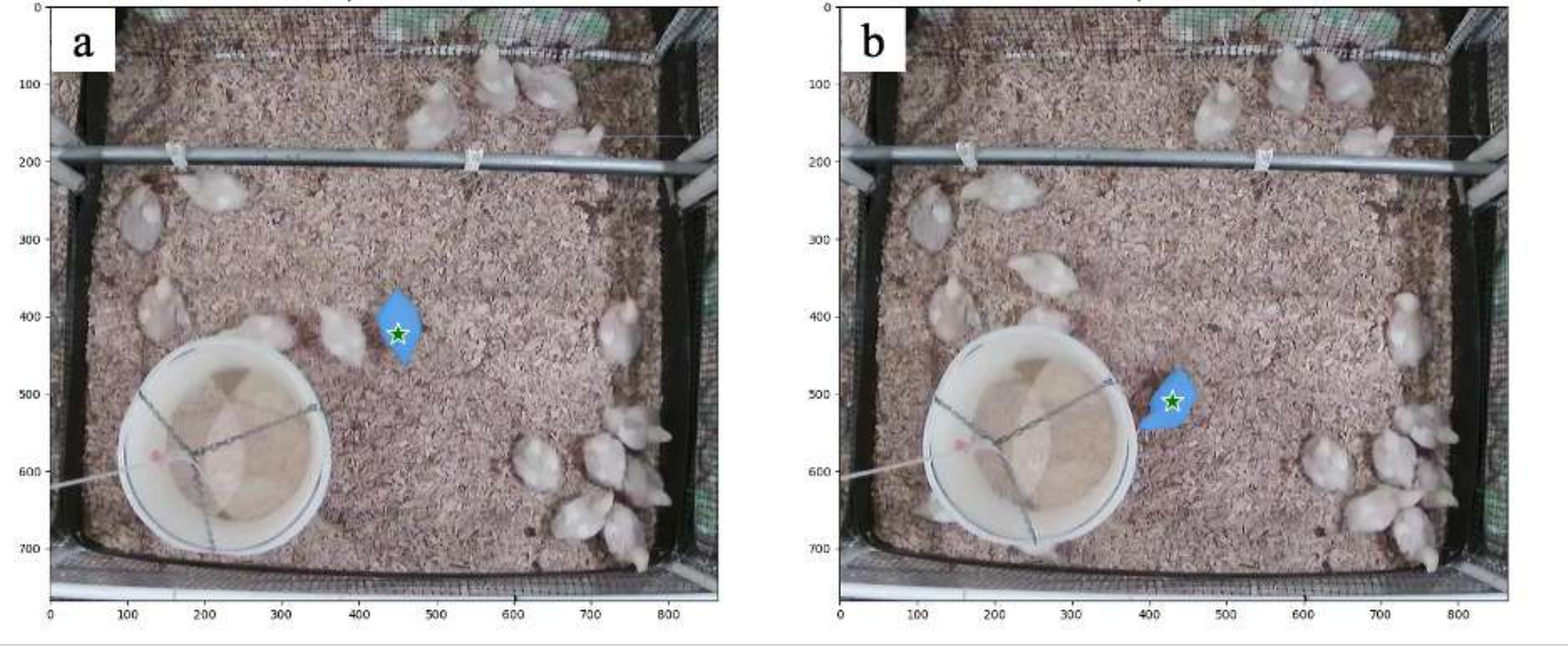}
    \caption{The tracking of a chicken using SAM+YOLOX+ByteTracker.}
    \label{fig:track}
\end{figure}
\subsection{Limitation}
Although SAM models have demonstrated remarkable performance in segmenting and tracking chickens, our evaluation has revealed several limitations that should be addressed in the future, including flock density, occlusion, and behaviors. SAM struggles with high flock densities where chickens have significant overlap. It is ineffective when chickens are obscured by structures such as feeders and nesting boxes. It may incorrectly detect chickens with altered postures or behaviors. 
We expect our paper can inspire future research to address these limitations and challenges and enhance the effectiveness and applicability of SAM in tracking and segmentation.

\begin{figure}[htbp]
\centering
\includegraphics[width=\linewidth]{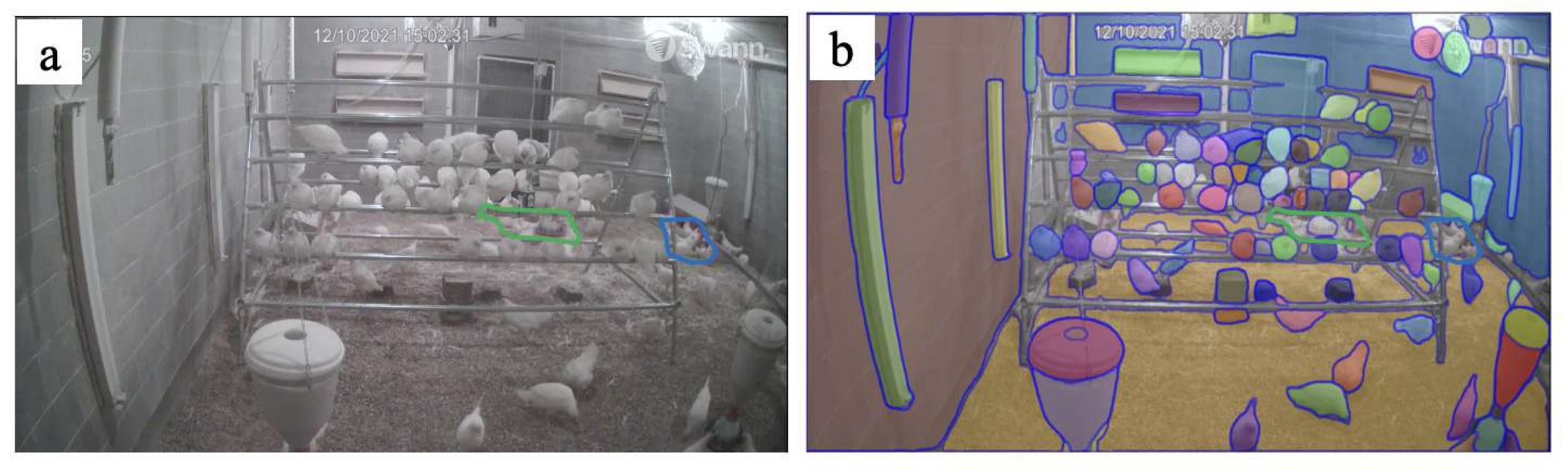}
\caption{Side view of chicken segmentation in a research cage-free house (a, b).}
\label{fig:pic3}
\end{figure}

\begin{figure}[htbp]
\centering
\includegraphics[width=\linewidth]{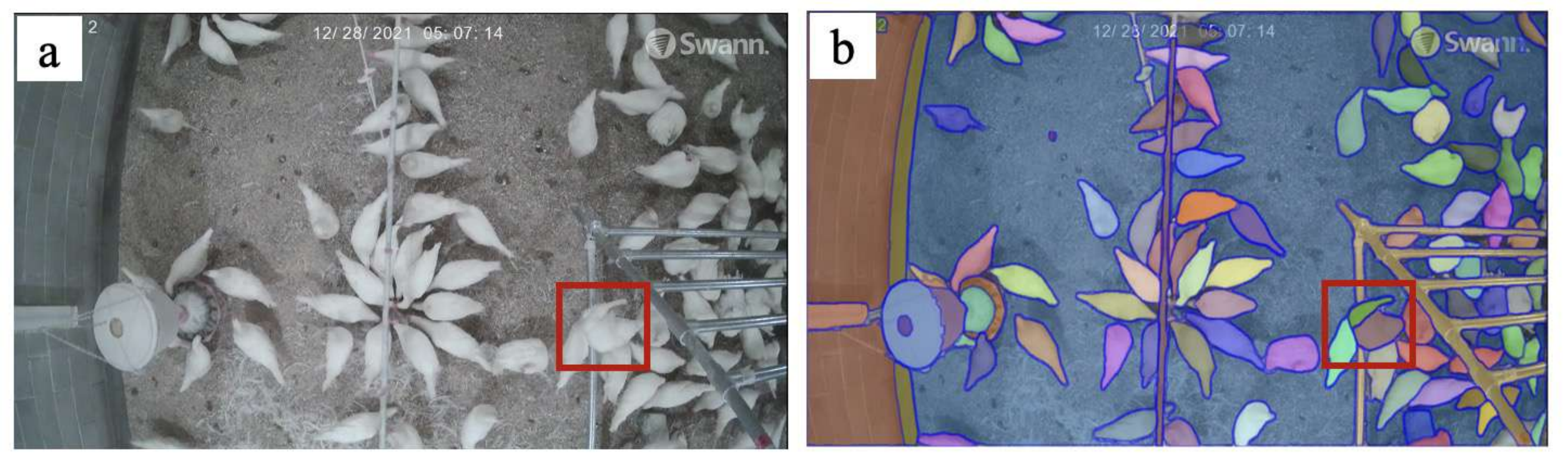}
\caption{Top view of chicken segmentation in a research cage-free house (a, b).}
\label{fig:pic4}
\end{figure}
\noindent\textbf{Flock density.} When flock density exceeds 9 birds/m\textsuperscript{2}, SAM struggles to segment individual chickens due to their overlapping bodies. This limitation is particularly evident in areas of high density, such as those circled in green in Fig. 3a and 3b. This results in detector errors and difficulty in accurately recognizing the hens' body shapes.

\noindent\textbf{Occlusion.} In cage-free housing systems, chickens can be occluded by various facilities such as drinkers, feeders, drinking lines, and nesting boxes, which can make it challenging for SAM to accurately detect them \cite{xue2023machine,subedi2023tracking}. For instance, in areas circled in blue in Fig. 3a and 3b, three hens are occluded by a perch frame, a drinking line, and a nesting box, respectively, resulting in missing detection of these hens.

\noindent\textbf{Behaviors.} When chickens exhibit different behaviors, their body postures change, leading to difficulty in accurately detecting them with SAM. For instance, in the red rectangle highlighted in Fig. 4a and 4b, a chicken is roosting with its head inside its feathers and its body more huddled in shape, which differs from a typical streamlined body shape. Without sufficient domain-specific training, SAM mistakenly detects the roosting chicken as two separate chickens.

\subsection{Future Work}
Given the remarkable segmentation outcomes achieved by SAM on the chicken body, further studies can investigate arbitrary parts of the chicken, such as legs, heads, and wings, and combine them with matched parts weights \cite{ichiura2019exploring, hnoohom2018thai}. This can enable the development of models like the wings weight prediction model, which can be used to sort wings based on their weight. Additionally, a bodyweight model can be created to monitor chickens' weight in real-time, enabling better control of the broiler time ready for market.
Moreover, combining the tracking function of SAM with more computer vision models can facilitate the monitoring of individual chicken behaviors, such as walking, drinking, and feeding \cite{lin2018monitoring}. By leveraging behavioral monitoring, a deeper understanding of chicken welfare can be achieved in a less invasive manner.
In the context of large foundation models, multimodal models can also be effectively deployed in the field of agriculture, with a particular emphasis on poultry segmentation. By providing fine-grained and targeted information, multimodal prompts could potentially enhance model performance, adapting to various poultry species, size variations, and diverse environmental conditions \cite{xiao2023instruction, wang2023seggpt}.

\section{Conclusions}
In this article, we discuss the development and evaluation of a new computer vision model, SAM, for chicken segmentation and tracking. The results showed that SAM outperformed state-of-the-art methods in both semantic and part-based segmentation, and the combination of SAM with YOLOX and ByteTracker allowed for real-time tracking of individual broiler bird movements. However, the study also revealed several limitations of SAM that could be addressed in future research, such as flock density, occlusion, and behaviors. The future work proposed in the article includes investigating arbitrary parts of the chicken, developing models for sorting and monitoring chicken weight, and combining SAM's tracking function with more computer vision models to monitor chicken behaviors. Overall, the article emphasizes the capacity of SAM to enhance chicken welfare and optimize poultry production operations.

\bibliographystyle{unsrt}
\bibliography{reference}

\begin{thebibliography}{10}

\bibitem{li2023learning}
Yu~Li, Junyi Yin, Shuoyan Liu, Bing Xue, Cyrus Shokoohi, Gang Ge, Menglei Hu,
  Tenghuan Li, Xue Tao, Zhi Rao, et~al.
\newblock Learning hand kinematics for parkinson's disease assessment using a
  multimodal sensor glove.
\newblock {\em Advanced Science}, page 2206982, 2023.

\bibitem{wang2021search}
Xiaolong Wang, Matthew Bilsky, and Subhrajit Bhattacharya.
\newblock Search-based configuration planning and motion control algorithms for
  a snake-like robot performing load-intensive operations.
\newblock {\em Autonomous Robots}, 45:1047--1076, 2021.

\bibitem{eli2019applications}
Ngozi~Clara Eli-Chukwu.
\newblock Applications of artificial intelligence in agriculture: A review.
\newblock {\em Engineering, Technology \& Applied Science Research},
  9(4):4377--4383, 2019.

\bibitem{lu2023agi}
Guoyu Lu, Sheng Li, Gengchen Mai, Jin Sun, Dajiang Zhu, Lilong Chai, Haijian
  Sun, Xianqiao Wang, Haixing Dai, Ninghao Liu, et~al.
\newblock Agi for agriculture.
\newblock {\em arXiv preprint arXiv:2304.06136}, 2023.

\bibitem{de2023semantic}
J~de~Curt{\`o}, I~de~Zarz{\`a}, and Carlos~T Calafate.
\newblock Semantic scene understanding with large language models on unmanned
  aerial vehicles.
\newblock {\em Drones}, 7(2):114, 2023.

\bibitem{stella2023can}
Francesco Stella, Cosimo Della~Santina, and Josie Hughes.
\newblock Can large language models design a robot?
\newblock {\em arXiv preprint arXiv:2303.15324}, 2023.

\bibitem{vaghefi2023chatipcc}
Saeid~Ashraf Vaghefi, Qian Wang, Veruska Muccione, Jingwei Ni, Mathias Kraus,
  Julia Bingler, Tobias Schimanski, Chiara Colesanti-Senni, Nicolas Webersinke,
  Christrian Huggel, et~al.
\newblock chatipcc: Grounding conversational ai in climate science.
\newblock {\em arXiv preprint arXiv:2304.05510}, 2023.

\bibitem{zhang2023self}
Dan Zhang and Fangfang Zhou.
\newblock Self-supervised image denoising for real-world images with
  context-aware transformer.
\newblock {\em IEEE Access}, 11:14340--14349, 2023.

\bibitem{luo2022multisource}
Xiaoling Luo, Xiaobo Ma, Matthew Munden, Yao-Jan Wu, and Yangsheng Jiang.
\newblock A multisource data approach for estimating vehicle queue length at
  metered on-ramps.
\newblock {\em Journal of Transportation Engineering, Part A: Systems},
  148(2):04021117, 2022.

\bibitem{kirillov2023segment}
Alexander Kirillov, Eric Mintun, Nikhila Ravi, Hanzi Mao, Chloe Rolland, Laura
  Gustafson, Tete Xiao, Spencer Whitehead, Alexander~C Berg, Wan-Yen Lo, et~al.
\newblock Segment anything.
\newblock {\em arXiv preprint arXiv:2304.02643}, 2023.

\bibitem{ji2023segment}
Wei Ji, Jingjing Li, Qi~Bi, Wenbo Li, and Li~Cheng.
\newblock Segment anything is not always perfect: An investigation of sam on
  different real-world applications.
\newblock {\em arXiv preprint arXiv:2304.05750}, 2023.

\bibitem{tang2023can}
Lv~Tang, Haoke Xiao, and Bo~Li.
\newblock Can sam segment anything? when sam meets camouflaged object
  detection.
\newblock {\em arXiv preprint arXiv:2304.04709}, 2023.

\bibitem{chen2023sam}
Tianrun Chen, Lanyun Zhu, Chaotao Ding, Runlong Cao, Shangzhan Zhang, Yan Wang,
  Papa Mao, and Ying Zang.
\newblock Sam fails to segment anything?--sam-adaptor: Adapting sam in
  underperformed scenes.
\newblock 2023.

\bibitem{subedi2023tracking}
Sachin Subedi, Ramesh Bist, Xiao Yang, and Lilong Chai.
\newblock Tracking pecking behaviors and damages of cage-free laying hens with
  machine vision technologies.
\newblock {\em Computers and Electronics in Agriculture}, 204:107545, 2023.

\bibitem{yang2022deep}
Xiao Yang, Lilong Chai, Ramesh~Bahadur Bist, Sachin Subedi, and Zihao Wu.
\newblock A deep learning model for detecting cage-free hens on the litter
  floor.
\newblock {\em Animals}, 12(15):1983, 2022.

\bibitem{yang2023deep}
Xiao Yang, Ramesh Bist, Sachin Subedi, and Lilong Chai.
\newblock A deep learning method for monitoring spatial distribution of
  cage-free hens.
\newblock {\em Artificial Intelligence in Agriculture}, 8:20--29, 2023.

\bibitem{yang2023automatic}
Xiao Yang, Ramesh Bist, Sachin Subedi, Zihao Wu, Tianming Liu, and Lilong Chai.
\newblock An automatic classifier for monitoring applied behaviors of cage-free
  laying hens with deep learning.
\newblock {\em Engineering Applications of Artificial Intelligence},
  123:106377, 2023.

\bibitem{bist2023mislaying}
Ramesh~Bahadur Bist, Xiao Yang, Sachin Subedi, and Lilong Chai.
\newblock Mislaying behavior detection in cage-free hens with deep learning
  technologies.
\newblock {\em Poultry Science}, page 102729, 2023.

\bibitem{ge2021yolox}
Zheng Ge, Songtao Liu, Feng Wang, Zeming Li, and Jian Sun.
\newblock Yolox: Exceeding yolo series in 2021.
\newblock {\em arXiv preprint arXiv:2107.08430}, 2021.

\bibitem{zhang2022bytetrack}
Yifu Zhang, Peize Sun, Yi~Jiang, Dongdong Yu, Fucheng Weng, Zehuan Yuan, Ping
  Luo, Wenyu Liu, and Xinggang Wang.
\newblock Bytetrack: Multi-object tracking by associating every detection box.
\newblock In {\em Computer Vision--ECCV 2022: 17th European Conference, Tel
  Aviv, Israel, October 23--27, 2022, Proceedings, Part XXII}, pages 1--21.
  Springer, 2022.

\bibitem{xie2021segformer}
Enze Xie, Wenhai Wang, Zhiding Yu, Anima Anandkumar, Jose~M Alvarez, and Ping
  Luo.
\newblock Segformer: Simple and efficient design for semantic segmentation with
  transformers.
\newblock {\em Advances in Neural Information Processing Systems},
  34:12077--12090, 2021.

\bibitem{liu2022setr}
Yang Liu, Yongfu Wang, Yang Li, Qiansheng Li, and Jian Wang.
\newblock Setr-yolov5n: A lightweight low-light lane curvature detection method
  based on fractional-order fusion model.
\newblock {\em IEEE Access}, 10:93003--93016, 2022.

\bibitem{resendiz2017segmentation}
Emmanuel Resendiz-Ochoa, Roque~A Osornio-Rios, Juan~P Benitez-Rangel, Luis~A
  Morales-Hernandez, and Rene de~J Romero-Troncoso.
\newblock Segmentation in thermography images for bearing defect analysis in
  induction motors.
\newblock In {\em 2017 IEEE 11th International Symposium on Diagnostics for
  Electrical Machines, Power Electronics and Drives (SDEMPED)}, pages 572--577.
  IEEE, 2017.

\bibitem{zhang2023feather}
Xiaomin Zhang, Yanning Zhang, Jinfeng Geng, Jinming Pan, Xinyao Huang, and
  Xiuqin Rao.
\newblock Feather damage monitoring system using rgb-depth-thermal model for
  chickens.
\newblock {\em Animals}, 13(1):126, 2023.

\bibitem{jing2023segment}
Yongcheng Jing, Xinchao Wang, and Dacheng Tao.
\newblock Segment anything in non-euclidean domains: Challenges and
  opportunities.
\newblock {\em arXiv preprint arXiv:2304.11595}, 2023.

\bibitem{chen2023learning}
Junzhang Chen and Xiangzhi Bai.
\newblock Learning to" segment anything" in thermal infrared images through
  knowledge distillation with a large scale dataset satir.
\newblock {\em arXiv preprint arXiv:2304.07969}, 2023.

\bibitem{ahmadi2023application}
Mohsen Ahmadi, Ahmad~Gholizadeh Lonbar, Abbas Sharifi, Ali~Tarlani Beris,
  Mohammadsadegh Nouri, and Amir~Sharifzadeh Javidi.
\newblock Application of segment anything model for civil infrastructure defect
  assessment.
\newblock {\em arXiv preprint arXiv:2304.12600}, 2023.

\bibitem{rong2023tomato}
Jiacheng Rong, Hui Zhou, Fan Zhang, Ting Yuan, and Pengbo Wang.
\newblock Tomato cluster detection and counting using improved yolov5 based on
  rgb-d fusion.
\newblock {\em Computers and Electronics in Agriculture}, 207:107741, 2023.

\bibitem{cao2023retinamot}
Jie Cao, Jianxun Zhang, Bowen Li, Linfeng Gao, and Jie Zhang.
\newblock Retinamot: rethinking anchor-free yolov5 for online multiple object
  tracking.
\newblock {\em Complex \& Intelligent Systems}, pages 1--19, 2023.

\bibitem{xue2023machine}
Hao Xue, Lihua Li, Peng Wen, and Meng Zhang.
\newblock A machine learning-based positioning method for poultry in cage
  environments.
\newblock {\em Computers and Electronics in Agriculture}, 208:107764, 2023.

\bibitem{ichiura2019exploring}
Shigeru Ichiura, Tomohiro Mori, KI~Horiguchi, and Mitsuhiko Katahira.
\newblock Exploring iot based broiler chicken management technology.
\newblock {\em Proceedings of the 7th TAE}, pages 205--211, 2019.

\bibitem{hnoohom2018thai}
Narit Hnoohom and Sumeth Yuenyong.
\newblock Thai fast food image classification using deep learning.
\newblock In {\em 2018 International ECTI northern section conference on
  electrical, electronics, computer and telecommunications engineering
  (ECTI-NCON)}, pages 116--119. IEEE, 2018.

\bibitem{lin2018monitoring}
Chen-Yi Lin, Kuang-Wen Hsieh, Yao-Chuan Tsai, and Yan-Fu Kuo.
\newblock Monitoring chicken heat stress using deep convolutional neural
  networks.
\newblock In {\em 2018 ASABE Annual International Meeting}, page~1. American
  Society of Agricultural and Biological Engineers, 2018.

\bibitem{xiao2023instruction}
Zhenxiang Xiao, Yuzhong Chen, Lu~Zhang, Junjie Yao, Zihao Wu, Xiaowei Yu,
  Yi~Pan, Lin Zhao, Chong Ma, Xinyu Liu, et~al.
\newblock Instruction-vit: Multi-modal prompts for instruction learning in vit.
\newblock {\em arXiv preprint arXiv:2305.00201}, 2023.

\bibitem{wang2023seggpt}
Xinlong Wang, Xiaosong Zhang, Yue Cao, Wen Wang, Chunhua Shen, and Tiejun
  Huang.
\newblock Seggpt: Segmenting everything in context.
\newblock {\em arXiv preprint arXiv:2304.03284}, 2023.

\end{thebibliography}

\end{document}